\documentclass[ai,article,accept,pdftex,moreauthors]{Definitions/mdpi} 
\usepackage{subcaption} % added<<<<<<<<<<<

\firstpage{1} 
\pubvolume{1}
\issuenum{1}
\articlenumber{0}
\pubyear{2025}
\copyrightyear{2025}
\externaleditor{Firstname Lastname} % More than 1 editor, please add `` and '' before the last editor name
\datereceived{ } 
\daterevised{ } % Comment out if no revised date
\dateaccepted{ } 
\datepublished{ } 
\hreflink{https://doi.org/} % If needed use \linebreak
\usepackage{subfig}
\usepackage{siunitx}

\Title{Self-Emotion-Mediated Exploration in Artificial Intelligence Mirrors: Findings from Cognitive Psychology} %MDPI: Title altered.

\TitleCitation{Self-Emotion-Mediated Exploration in Artificial Intelligence Mirrors: Findings from Cognitive Psychology}

\Author{Gustavo Assuncao $^{1,3,}$*\orcidA{}, %MDPI: 1. Please carefully check the accuracy of names and affiliations; 2. We added ``*'' after author name according to the system, please check; 3. The order of last two author name different from susy, please check which one is correct
 Miguel Castelo-Branco $^{1,2}$\orcidB{} and Paulo Menezes $^{1,3}$\orcidC{}}

\AuthorNames{Gustavo Assuncao, Miguel Castelo-Branco and Paulo Menezes}
\AuthorCitation{Assuncao, G.; Castelo-Branco, M.; Menezes, P.}
\address{%
$^{1}$ \quad University of Coimbra, %MDPI: The department/school/faculty/campus of this university is required. Please try to provide this information.
 Department of Electrical and Computer Engineering, 3030-790 Coimbra, Portugal; mcbranco@fmed.uc.pt (M.C.B.); pm@deec.uc.pt~(P.M.) %MDPI: We added these email addresses here according to those submitted online at susy.mdpi.com. Please confirm.
\\
$^{2}$ \quad Institute for Biomedical Imaging and Translational Research (CIBIT), ICNAS, 3000-548 Coimbra, Portugal \\
$^{3}$ \quad Institute of Systems and Robotics (ISR), DEEC, 3030-290 Coimbra,  Portugal
}

\corres{Correspondence: gustavo.assuncao@isr.uc.pt}

\abstract{Background: Exploration of the physical environment is an indispensable precursor to information acquisition and knowledge consolidation for living organisms. Yet, current artificial intelligence models lack these autonomy capabilities during training, hindering their adaptability. This work proposes a learning framework for artificial agents to obtain an intrinsic exploratory drive, based on epistemic and achievement emotions triggered during data observation. Methods: This study proposes a dual-module reinforcement framework, where data analysis scores dictate {pride or surprise,} in accordance with psychological studies on humans. A correlation between these states and exploration is then optimized for agents to meet their learning goals. Results: Causal relationships between states and exploration are demonstrated by the majority of agents. {A $15.4\%$ mean increase is noted for surprise, with a $2.8\%$ mean decrease for pride. Resulting correlations of $\rho_{surprise} = 0.461$ and $\rho_{pride} = -0.237$ are obtained,} mirroring previously reported human behavior. Conclusions: These findings {lead to the conclusion that bio-inspiration for AI development can be of great use. This can incur benefits typically found in living beings, such as autonomy. Further, it empirically shows how AI methodologies can corroborate human behavioral findings, showcasing major interdisciplinary importance.} Ramifications are discussed.
}

\keyword{exploration, artificial emotion, general artificial intelligence, reinforcement learning, intrinsic drives} 

\begin{document}

\section{Introduction}
Recent advances in AI have led to a surpassing of conventional methodologies and disruption of various human-powered fields. These are becoming more and more digital, from~medical pathology~\cite{Baxi_2021} to industrial smart systems~\cite{Sarker_2022} and even socio-emotive companionship~\cite{Assuncao_2022}. However, this prosperity is fickle as the existing AI methodology is largely ineffective {when} devoid of human guidance~\cite{M_hlhoff_2019}. This is a clear indicator that research on AI learning should move closer to metalearning. Additionally, the parameter explicitness required of model designers should be considered when building training procedures for artificial agents (naturally without intrinsic motivation). In~this context, we postulate artificial emotion as a missing catalyst of exploratory behavior in AI and~develop primary work on how to use it for that goal. This particular characteristic {can} enable agents to focus on data in accordance with their needs/interests, given how appraisal and attention changes correlate to alter sensory processing of stimuli in~an effect dubbed emotional salience~\cite{Pauli_2008}. In~fact, exploration is already known to be a fundamental aspect of cognitive development and independent behavior in human beings~\cite{Nunnally1974} given its contribution to the acquisition of knowledge. {For instance,} confirmation biases influence how information is sought to~ratify prior beliefs and inference~\cite{Kaanders_2022}. Consequently, {for} AI to grow autonomous, {researchers} should strive to develop methodologies congruent with {biological processing} and optimize informational~search.

Scrutiny of epistemic/achievement states (i.e., emotions pertaining to the generation of knowledge and a sense of success) for the purpose of benefiting exploration tactics in AI has been attempted, {though only} from {a few general} perspectives. Approaches have applied model behavior differences as criteria to determine whether data is adequate for classification~\cite{8812069, Weiss_2021}, and~divergence in transition probability has been employed as learning reinforcement~\cite{Achiam2017}. There are also instances of states being inferred from intrinsic reward~\cite{Yin_2021} to~drive curiosity in exploration. Results {were} interpreted as congruent with emotion in real life, {where} internal cognitive conditions {related with emotion (e.g.,} incongruity and expectancy{) do influence} exploration~\cite{Marshall_2006, Pekrun2019}. However, {these interpretations only posit conditions as} causes of emotional variation, which then impacts exploration. {Hence}, emotion is only implicit and does {not benefit from the better contextual adaptability, informational density, or~socio-communicative relevance that explicitness can bring. This leads to} low generalizability. {Practical applications are lost, such as the impact over learning aspects besides exploration, human understanding of AI decision-making, or~the ability to further integrate environmental information.} {Contrastingly,} emotional influence is well acknowledged for {several} biological factors~\cite{Damasio_1994}. {Thus,} reproducing conditions for direct emotion manifestation and {consequential} impact may represent a better methodology {than current state-of-the-art approaches}.

This {improvement} is corroborated by works which have achieved greater performances from autonomous emotion-mediated parameter optimization. For~{example}, works on the prediction error, learning rate, or actual reward have presented emotional modulation based on the difference between short- and long-term average of reward entropy over time~\cite{huang_2018} or~emotion quantization as a linear combination of separate power levels~\cite{wang_2021}. %MDPI: Please check that the intended meaning has been retained.
A mere difference in visual stimuli has also been made to influence valence--arousal pairs~\cite{hieida_2018, hieida2_2018}, {which then} affect learning parameters. Moreover, the~basing of emotion metalearning techniques on neurophysiology, where researchers strive to replicate limbic circuitry and neuromodulation, has been shown to provide performance advantages {in decision-making}~\cite{Assuncao_2022}. It is a~{type of} strategy that benefits from interdisciplinarity, {yet it is uncommon}. {As such,} it could provide an edge when developing new parameter calculation techniques and contribute to {emotion-mediated AI progress}. 

{Linking} the lack of learning autonomy in artificial agents with an observable influence of intrinsic {emotional} drives in living beings, there is motivation to emulate the latter {in an attempt} to mitigate the former. This emulation should build on knowledge already established by neuropsychological studies as~a bootstrapping point. Furthermore, it should remain task-agnostic yet still provide some type of advantage for agent learning, either in terms of autonomy or efficiency. {In order to} tackle the challenges of endowing AI with {emotion-mediated} intrinsic driving {and study its outcomes}, this paper considers the following two research~questions:
\begin{itemize}
    \item RQ 1 %MDPI: Please confirm if the italics are necessary; if not, please remove them. The following highlights are the same.
: How can emotion be represented in artificial agents, and~how will its influence over their behavior be evaluated so that results may be valid and comparable to human behavioral studies? To achieve this, we {first build epistemic and achievement emotion functions based on links demonstrated in cognitive psychology. These are applied to a learning framework, whose behavior is evaluated under conditions similar to those of human studies}~\cite{Chevrier_2019, Vogl_2019, Vogl_2020, slinger2022}.
    \item RQ 2: Will the manifested correlations between emotion and exploratory behavior be useful {for data processing by agents}, similar to what happens with human {beings}? {To understand this we integrate} the framework in a learning loop, replicated over a large number of agents, {and assess emergent correlations.} Parallelism {is then drawn} between the former and reported human behavior. 
\end{itemize}
{Answering these questions is meant to contribute a valid technique for artificial agents to explore data more autonomously. It is also meant to spark more interest in human--AI interdisciplinary studies.} The rest of the paper is organized as follows: Section~2 briefly introduces the topic of emotion--behavior studies in psychology after~framing our approach within the context of AI. Section~3 describes the design of each framework component, subsequently overviewing the experimental arrangement inspired by human studies. Section~4 details the experimental procedure and obtained results, followed by Section~5, which discusses them. Finally, Section~6 concludes the~paper.

%In the present work we propose a framework for determining appropriate exploration ratios in agents performing a task, based on which the exploratory-triggering emotion is derived, motivated by extensive observation of links connecting separate epistemic and achievement emotional states to exploratory behavior in cognitive psychology~\cite{Chevrier_2019, Vogl_2019, Vogl_2020, slinger2022}. 
%The next section will describe the various stages of development of this framework. First, the construction of internal state functions will be specified, based on emotional variation literature. Subsequently, deep learning methodology will be justified as loosely emulating the neurophysiology responsible for emotion-induced attention mechanisms and voluntary action. The objective here is to enable AI with overt emotion to employ it as an exploratory drive, assessing performance to determine usability. Moreover, this is intended in order to determine the level of similarity with the human data motivating our design~\cite{Vogl_2019, Vogl_2020}, fomenting interdisciplinarity. Specifically, our framework and experimental setting corroborates observations of human behavior, providing a basis for further collaborative research. The rest of the paper is organized as follows: 

\section{Background and Related~Work}
This section overviews the neuropsychological motivation behind {our approach and~its importance for interdisciplinary analogy.} It further presents works tackling learning {mediation by emotion} in artificial intelligence, framing {its usefulness and exposing the novelty of this work}.

    \subsection{From Psychology to~AI}
    The study of links connecting epistemic and achievement emotions (i.e., {states} pertaining to the generation of knowledge and a personal sense of success) to exploratory behavior has been an active {topic} of research in cognitive psychology~\cite{Chevrier_2019, Vogl_2019, Vogl_2020, slinger2022}. {In behavioral studies}, epistemic emotions are commonly observed transitioning into confused and curious demeanors~\cite{Muis_2018}. They also potentially lead to heightened motivation and the pursuit of success~\cite{sznycer2021}. {Depending on its outcome, the~pursuit can lead} to pride or shame. {If} confronted with information contradictory of internalized knowledge {(i.e., high-confidence errors)}, people {can} manifest surprise supplanted by unexpected error outcomes. {All in all,} experiencing a positive outcome in a task predisposes humans into seeking similar internal reactions {and corresponding scenarios}. {This exploratory behavior happens for both epistemic and achievement} cognitive paths~\cite{Vogl_2019}, albeit with potentially different objectives. {The process is regulated by the brain's reward system, which adapts its} signaling proportionally to the aforementioned conditions~\cite{Schultz_2015}. Considering those are reproducible {in} deep learning, epistemic and achievement emotions {may be integrated in AI for~exploratory benefits.}

    {For studying} the effects of emotion over human behavior, the methodology is straightforward. {It typically} relies on tasks~\cite{HACKMAN196997}, such as classification and trivia, {designed} to induce {relevant} scenarios. Specifically in the works by Vogl~et~al.~\cite{Vogl_2019, Vogl_2020}, adulteration of common knowledge statements is presented in a veracity assessment task. {These} statements induce errors and {trigger} epistemic/achievement states, {which translate into exploration. For~instance,} uninformed human participants {being} presented incorrect statements {became surprised, as} confidence on their personal knowledge {clashed} with mistakes. Complementarily, correct responses prompted a sense of pride. {Either scenario sparked} exploration, as demonstrated by requests for additional~information. 
    
    {AI-oriented} tasks are {evidently} different from those humans perform {in} experiments. {Regardless,} this does not invalidate adaptation to a machine friendlier format, so comparable observations may be taken~\cite{Korteling_2021}. {This is corroborated by the fact that AI} is considered a valid framework on which to explore a range of {neuropsychologic} phenomena~\cite{Dawson_2014, Dawson_2009}. {Our contribution adds to this by~adapting Vogl~et~al. experimental conditions for an AI application. Confident participants can be represented by models with} near-perfect performance. {Cognitive tasks can be directly emulated, for~example, as} simple classifications. {Naturally, so can knowledge adulteration, which enables emotional triggering. With~this notion, the~problem became a matter of adequately integrating emotion in artificial agents. Then these can be validly} regarded as participants in a cognitive psychology~experiment.

    \subsection{Emotion-Driven~Learning}
    {Most} studies {integrating} emotion in AI {rarely draw from} robust psychological and neurophysiological {foundations}. Fewer employ it as a driver of learning. {Still}, our work relates with {some} studies. {For instance, in}~\cite{Mazzaglia2022}, the {authors} developed a latent dynamics model by {calculating} the dissimilarity from its posterior to prior beliefs. {This} rewarded exploration when it occurred. Schillaci~et~al. also presented a process for estimating the change in prediction error (PE) as a metric of learning progress~\cite{Schillaci_2020}. This enabled agents to shift attention towards more interesting (change-inducing) goals when progress {was} inadequate. In~\cite{JMLR:v23:21-0808}, the authors suggest a form of intrinsic rewarding reliant on competence progress, which is analogous to achievement gratification. {Based on it,} agents can {decide how to} explore their goal space. While {these approaches advance autonomous exploration, they overlook its intrinsic driving background, such as emotion. This omission limits contextual relevance and narrows the applicability of the resulting metrics}~\cite{Assuncao_2022}{. They also disregard most parallelism with living systems, despite their objective usefulness for understanding exploratory origins. Besides~potentially slowing AI advancement, this also incapacitates interdisciplinary understanding}, since inspiration from biological functioning {can help} postulate novel theories on how cognition develops from basic neural activity~\cite{Storrs2019DeepLF}. 

    Our {contribution is two-fold, as~the work} bears considerable interdisciplinary interest {besides} being advantageous towards autonomous AI. {First, we propose building agents} in accordance with neuronal structuring. {This novelism can mediate exploration, without~the hindrances of related work. It also increases the plausibility that resulting} exploratory tendencies are useful for understanding {real} neuronal arrangements~\cite{Macpherson_2021}. {Second, we further innovate by adapting the} experimental conditions under which psychological studies assess emotion--exploration relationships in humans { for~AI agents. This is more akin to the trend of}~\cite{Piloto_2022}, where psychological findings were considered as a basis for designing AI experiments. Despite the necessary adaptations for a standard AI methodology, the~resulting framework {can adequately demonstrate that autonomous exploration is possible. Moreover, it can appropriately} corroborate {psychological} findings and~provide a basis for other hypotheses to be considered, which are otherwise not easily {achieved in behavioral studies}.

    \subsection[Ethics]{{Ethics}}
    {Integrating emotion in AI systems introduces a layer of behavioral modulation that aligns with human interaction. This can have several benefits, as~specified, but~also raises important ethical concerns. One such concern is unpredictable behavior. As~decision-making ties with emotional metrics, optimization of underlying correlations may deviate from what is safe into what agents need for success. In~a controlled experimental environment, this is unlikely to cause harm. However, in~the real world, there may be risks to human--agent interactions and potential boundary violation, social or~physical. 

    The anthropomorphization of AI with emotion can also blur distinctions between real and simulated feeling. This can entail ascription of trust and compliance with systems liable to external manipulation. This can challenge broader social norms around responsibility and accountability, as~it becomes unclear whether the system, designer, or~improper manipulator should be held accountable for decision outcomes. While ethical concerns are valid, emotion-driven AI offers clear benefits when used responsibly. Emotions can serve as functional signals that enhance adaptability, learning, and~human interactions, provided systems are designed with transparency. For~an in-depth overview of AI-related ethics, readers are invited to check}~\cite{9844014}.

    In summary, {our} paradigm was expected to demonstrate a causal relationship where epistemic {or} achievement emotions served as mediators of exploratory behavior, mimicking the findings reported by Vogl~et~al.~\cite{Vogl_2019}. {Novelism stems from how the proposed framework is built, as~well as from the comparison of emergent behaviors with those of humans. Finally,} the resulting impact over agent knowledge acquisition and~overall behavior {was also considered useful contributions for AI} autonomy.

%%%%%%%%%%%%%%%%%%%%%%%%%%%%%%%%%%%%%%%%%%
\section{Materials and~Methods}
\begin{figure}[H]
        \centering
        {\includegraphics[width=0.95\columnwidth]{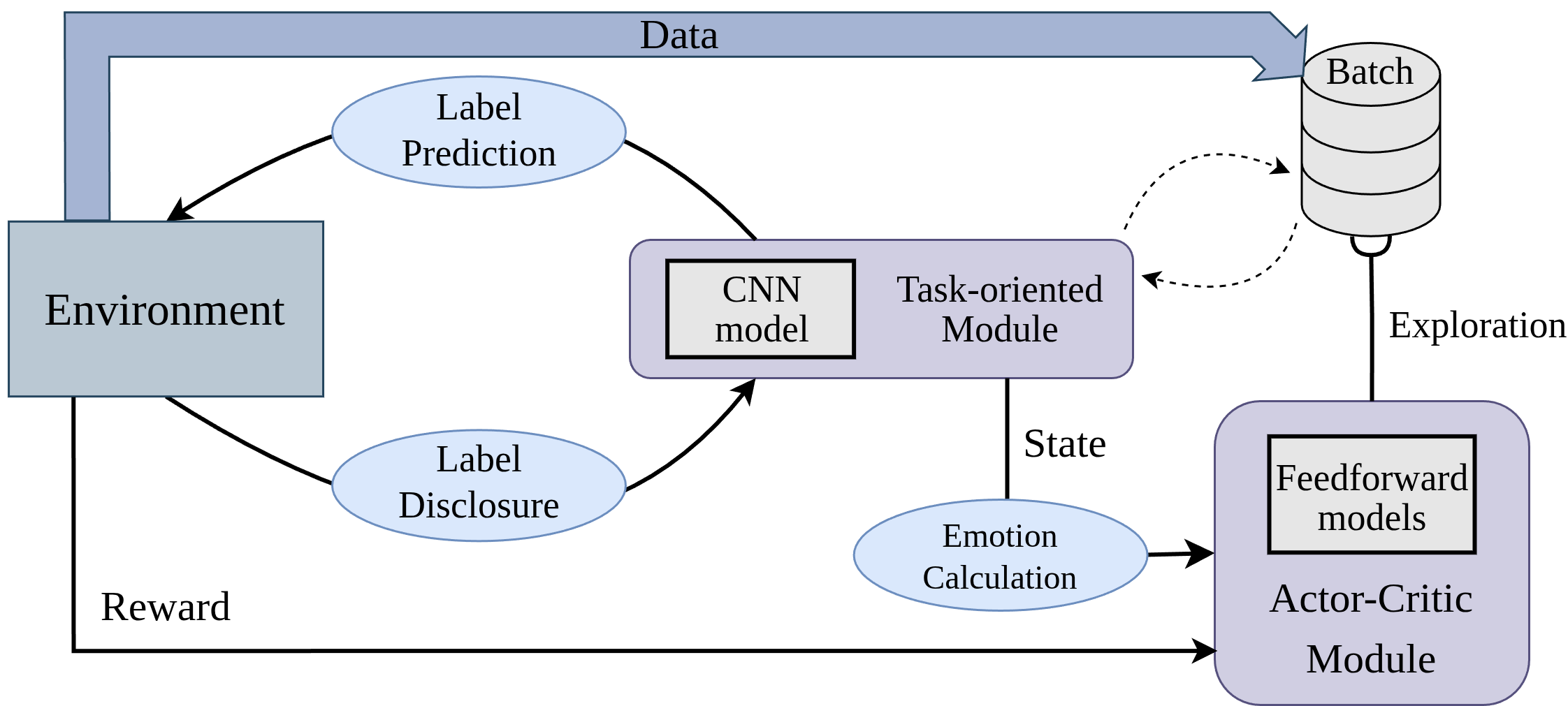}}
        \caption{{Framework overview, demonstrating how its components interact with the environment and compute an exploratory ration from performance-based emotion.}}
        \label{Fig:overview}
    \end{figure}

The proposed framework consists of a task-oriented module, whose rate of exploration is dictated by an actor--critic module. {The latter derives this rate} from performance-based emotional scoring. The following sections overview each component of {the} system, namely how the emotional functions were replicated from psychology observations, what composes the framework, and~{how} the learning cycle is {designed}. Figure~\ref{Fig:overview} displays the proposed framework as~a reference.

\unskip

    \subsection{Replication of Surprise and~Pride}
    We propose {deriving} epistemic and achievement emotion {from} the standard performance metrics of a {deep learning methodology, interpreting them as underlying cognitive conditions.} Specifically, testing accuracy reflects the adequacy of a model towards some task by~gauging overall correctness over unseen data. Therefore, it may be employed as a pointer of error and achievement. In~this {case}, escalation in accuracy scores can be interpreted as increasing success. {Contrastingly,} de-escalation entails a decrease in {success}. Variations in the feeling of pride should therefore match variations in accuracy, corresponding to personal achievement or lack thereof~\cite{Utz_2018}. {This} accuracy--pride match can entail a curve with a positive slope and unknown convexity, with~small variations. Factoring in confidence, {besides} accuracy, can broaden the set of representable emotions. {For instance}, high-confidence errors trigger the feeling of surprise, as {derived from} the cognitive incongruity {explained} previously. {Additionally}, insecure attainment of success may also induce surprise~\cite{Gendolla_1997}. In~these scenarios, {a respective} decrease or increase in confidence will instead lower surprise. A~saddle-like behavior can therefore describe this emotion, as~polarized variations of accuracy and confidence together imply intense values of surprise, whilst matching magnitudes of the two indicate a reduction or lack {of this emotion}. This view regarding surprise and pride is widely backed by the cognitive psychology literature~\cite{Vogl_2019, Vogl_2020, Marshall_2006, Pekrun2019}.

\begin{figure}[H]

%\begin{adjustwidth}{-\extralength}{0cm}
%\centering %% If there is a figure in wide page, please release command \centering, for Table, ``\textwidth" should be ``\fulllength"
\begin{tabular}{cc}
       %\centering
     % \subfloat[Pride\label{fig:pride}]{
        \includegraphics[width=0.45\linewidth]{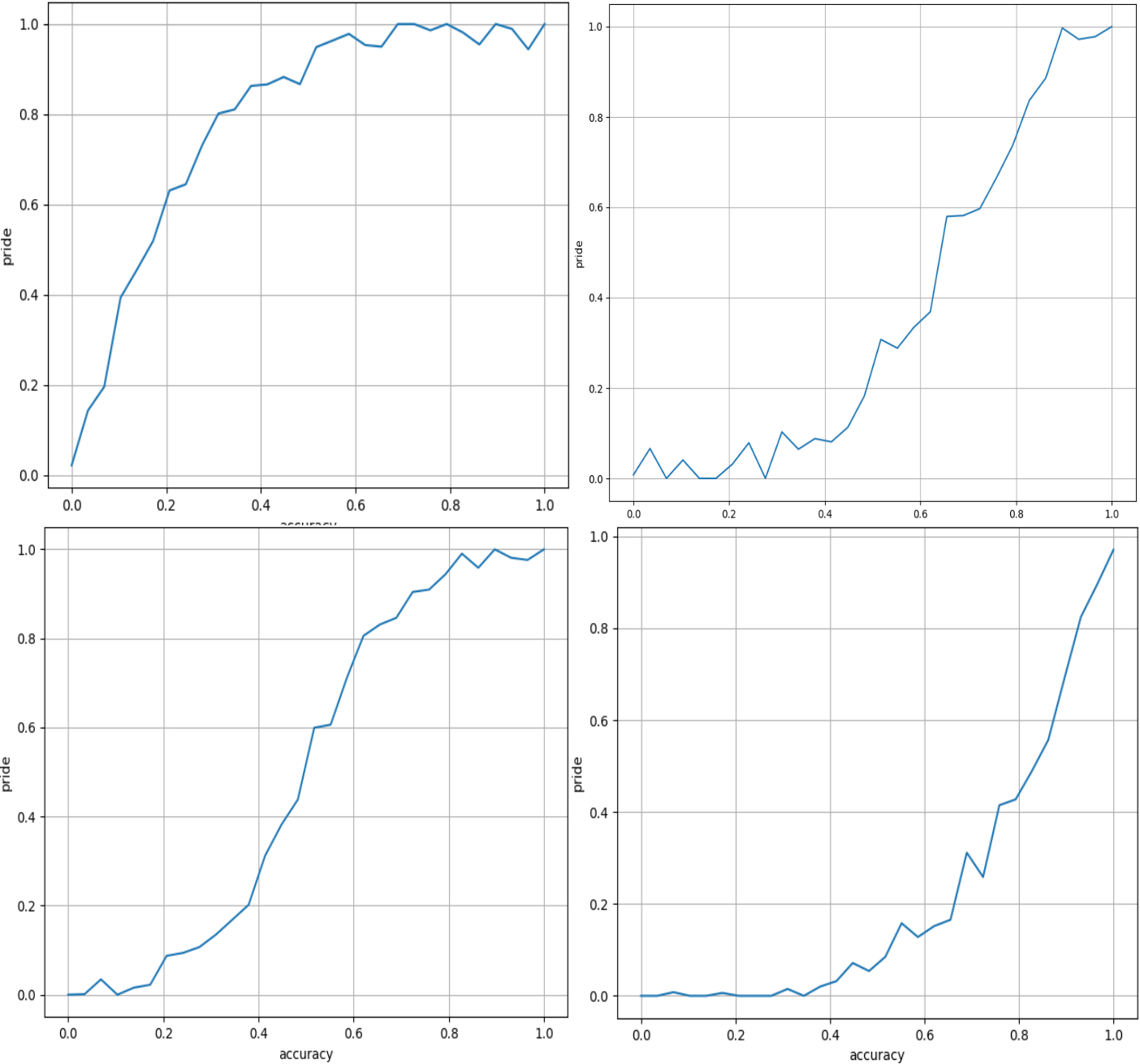}&
        %} \hfill
    %  \subfloat[Surprise\label{fig:surprise}]{
        \includegraphics[width=0.45\linewidth]{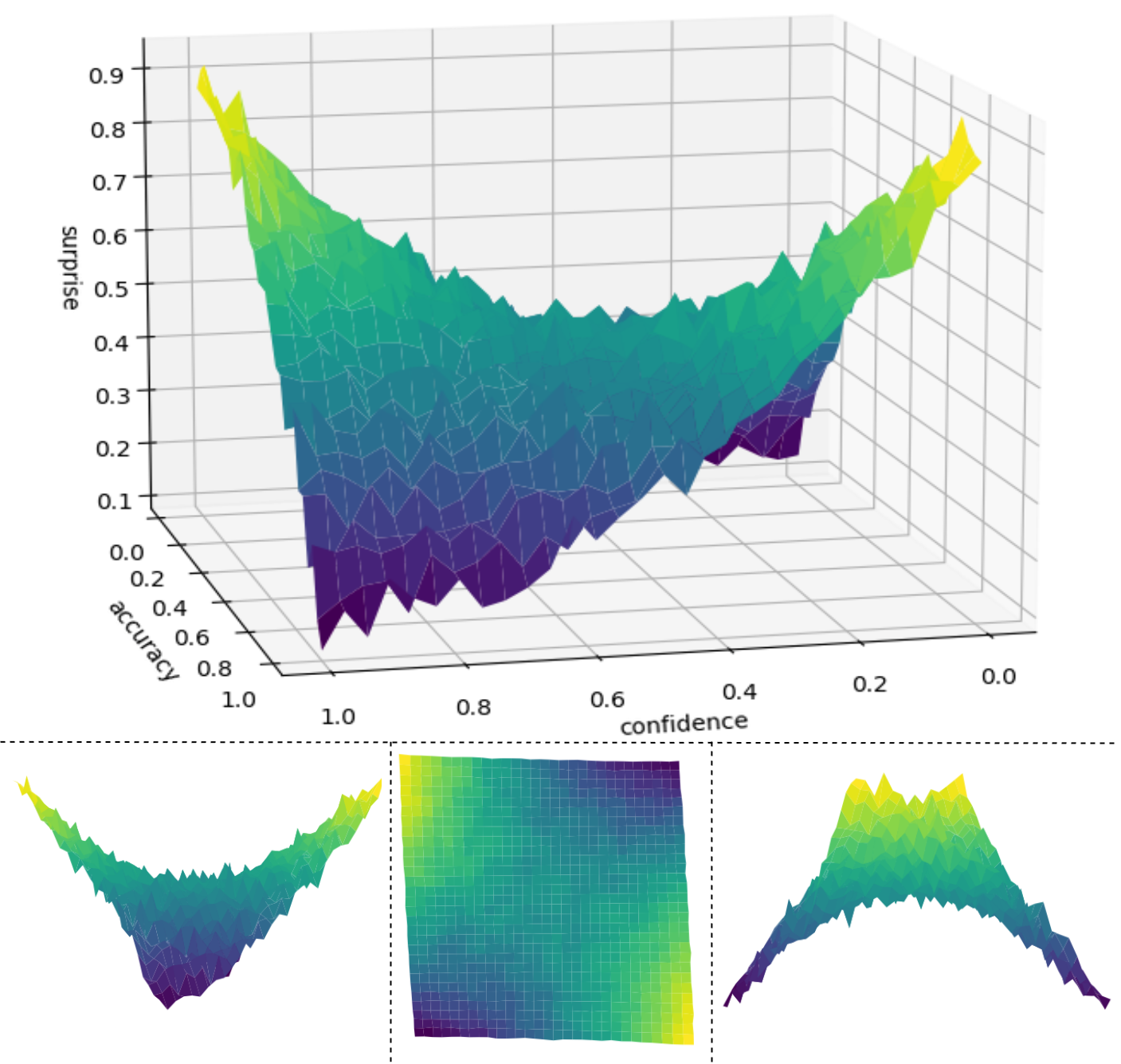}\\
({\bf a}) Pride &({\bf b}) Surprise\\
      %  }
\end{tabular}
%\end{adjustwidth}

      \caption{{Example artificial emotion curves inspired by cognitive psychology }\cite{Vogl_2019, Vogl_2020}{. (\textbf{a}) Positive pride slope based on increasing accuracy. (\textbf{b}) Saddle-like behavior of surprise based on accuracy and confidence.}}
      \label{Fig:emotions}
    \end{figure}

    {Several different functions can represent the behavior described for surprise and pride. Here,} a single set of functions {that fulfill the requirements} was selected arbitrarily for the main experiments. {Examples are shown in Figure }\ref{Fig:emotions} {for~reference.} As described, these examples factored in performance metrics of the task-oriented module, which is {inspired by} the impact of action outcomes over emotion~\cite{Kiuru_2020}. {This partially answers RQ1, contributing a novel way for emotion to occur in artificial agents, making it both objective and integrable in other AI~pipelines.}

    \subsubsection[Pride]{{Pride}}
    %Considering accuracy has been observed to positively predict pride~\cite{Vogl_2019}, in addition to being readily available and already bound to a $[0,1]$ interval, the variation of this achievement emotion $P$ over increasing accuracy $a$ would likely entail a curve of positive slope and unknown convexity, with occasional fluctuations. Thus, a possible example of the pride function could be:
    {Since accuracy has been shown to positively predict pride}~\cite{Vogl_2019}{, making it is easy to compute and already fixed within $[0,1]$, it can adequately model this emotion. To~capture how pride might vary with increasing accuracy $a$, we propose a Gaussian-like bump function $P(a)$. This displays an overall upward trend, peaking near $a = 1$ (perfect accuracy), to~reflect strong pride near high performance. We also include minor fluctuations to account for individual variability (e.g., personality and context). An~example of such a function is}
\begin{align}
    \begin{split}
        P \colon [0,1] &\to [0,1]\\
        a &\mapsto Clip\left[({100 \cdot C_1)}^{-(a-1)^2} + \mathcal{N}(\mu, \sigma^2)\right] 
    \end{split} 
    \tag{{$1$}} \label{eq:1}
    \end{align} 

    {In the equation, $C_1 > 1$ controls how sharply pride rises as accuracy improves. The~Gaussian noise $\mathcal{N}(\mu, \sigma^2)$ introduces individual variability. The~clip function keeps pride within its natural bounds (0 and 1). This example is not meant to be precise or universal, but~rather to illustrate a plausible emotional trajectory. Pride grows with success, but~its exact path varies across individuals.}
    
    \subsubsection[Surprise]{{Surprise}}
    %On a separate note, surprise is positively predicted by high-confidence errors~\cite{Vogl_2019}, meaning another metric is necessary besides accuracy to represent its behavior. Thus, a confidence score $c$ bound to the interval $[0.8,1]$ was introduced to simulate the high levels of confidence expected of a newly trained well-performing task-oriented module. The saddle-like rough surface of this epistemic emotion $S$ could therefore be obtained from:

    {Surprise depends on confidence, besides~accuracy. Prior work shows that high-confidence errors are strong predictors of surprise}~\cite{Vogl_2019}{. To~reflect this, we introduce a confidence score $c \in [0.8, 1]$, capturing the elevated confidence distribution yielded by the task-oriented model. Our surprise function $S(c,a)$ is proposed to provide high values when confidence and accuracy disagree (saddle-like behavior). This happens when an agent is confident but wrong (high $c$ and low $a$) or~unsure but correct (low $c$ and high $a$). It takes the following form:}
\begin{align}
    \begin{split}
        S \colon [0,&1]^2 \to [0,1]\\
        c,a &\mapsto Clip\left[ \mathcal{T}\left(\mathcal{R}\left( a^2 - c^2 \right)\right) + 0.5 + \mathcal{N}(\mu, \sigma^2)\right]
    \end{split} 
    \tag{{$2$}} \label{eq:2}
    \end{align}

    {Here, the~core term $a^2 - c^2$ captures the disagreement between accuracy and confidence. A~rotation $\mathcal{R}$ (by $45^{\circ} \pm C_2$) and translation $\mathcal{T}$ are applied to center and orient the surface so that surprise is maximized when accuracy and confidence diverge. As~before, Gaussian noise introduces variability, and~clipping keeps outputs within the range $[0, 1]$. The~$0.5$ term re-centers the output toward the range middle, reducing clipping distortion.}

    \subsection[Framework Overview]{Framework {Overview}}
    In the framework, three models were implemented: the task-oriented, actor, and critic models. This combination was loosely inspired by human neural functioning during a psychological experiment, {where part of the brain is focused on task success (e.g., prefrontal cortex), which is mediated by feedback from other regions (e.g., basal ganglia)}. It constituted each artificial participant, being fed data for classification with potentially wrong labels. {This biological parallelism was purposeful so that the obtained results could be contrasted with behavioral studies with adequate validity, in~line with answering RQ1 and fomenting human--AI interdisciplinarity in state-of-the-art research. %MDPI: Please check that the intended meaning has been retained.
   Furthermore, this clear separation between the task-oriented module and emotion to exploration optimization expands on current AI exploration techniques, which usually combine them and disregard bio-inspiration.}

    \subsubsection{Task-Oriented~Module}
    This module is meant to carry out a cognitive task. Here, handwritten digit recognition was performed using the MNIST dataset~\cite{deng2012mnist} for~the sake of simplicity. Other tasks would also be possible, as~the module is task-agnostic. Convolutional and feedforward branches were combined in a VGG-like architecture~\cite{VGG} to~first enhance visual cues representative of the content in images and then classify them {as digits}. The~architecture is simple, with~the layer arrangement outlined by Table \ref{tab:my-table}.
    
    \begin{table}[H]
      %  \centering
        \caption{{Task-oriented model architecture.}}
        \label{tab:my-table}
       % \noindent\colorbox{yellow}{
       % \resizebox{0.85\columnwidth}{!}{%
        \begin{tabularx}{\textwidth}{ccccc}
        \toprule
        \textbf{Layer} & \textbf{Type} & \textbf{Kernel/Units} & \textbf{Activation} & \textbf{Output Shape} \\ \midrule
        1 & Conv2D     & 3 $\times$ 3 (32 filters) & ReLU    & 26 $\times$ 26 $\times$ 32 \\
        2 & MaxPooling & 2 $\times$ 2              & -       & 13 $\times$ 13 $\times$ 32 \\
        3 & Conv2D     & 3 $\times$ 3 (64 filters) & ReLU    & 11 $\times$ 11 $\times$ 64 \\
        4 & MaxPooling & 2 $\times$ 2              & -       & 5 $\times$ 5 $\times$ 64   \\
        5 & Flatten    & -                  & -       & 1600         \\
        6 & Dropout    & 0.5                & -       & 1600         \\
        7 & Dense      & 128                & ReLU    & 128          \\
        8 & Dense      & 10                 & Softmax & 10     \\
        \bottomrule     
        \end{tabularx}%
       % }
      %  }
    \end{table}
    
    In order to train this model, half of the MNIST {training} dataset was used unadulterated. Standard backpropagation was employed using the Adam optimizer~\cite{adam} for 50 epochs, with~a batch size of 64. {Testing with the MNIST test set yielded} over $99\%$ accuracy and near-zero loss. {The confidence distribution is heavily concentrated in the high-confidence bins, with~over 98\% of samples falling within the [0.95, 1] range. This indicates that the model is highly confident in its predictions. Consequently, it is well-suited for applications in our framework, where such consistent certainty parallels the behavior of a person that is highly confident as~a result of repeated success. This further validates an answer to RQ1.}

\begin{figure}[H]
\begin{tabular}{cc}
      %\centering
     % \subfloat[Task-oriented module\label{fig:task_oriented}]{
        \includegraphics[width=0.6\linewidth]{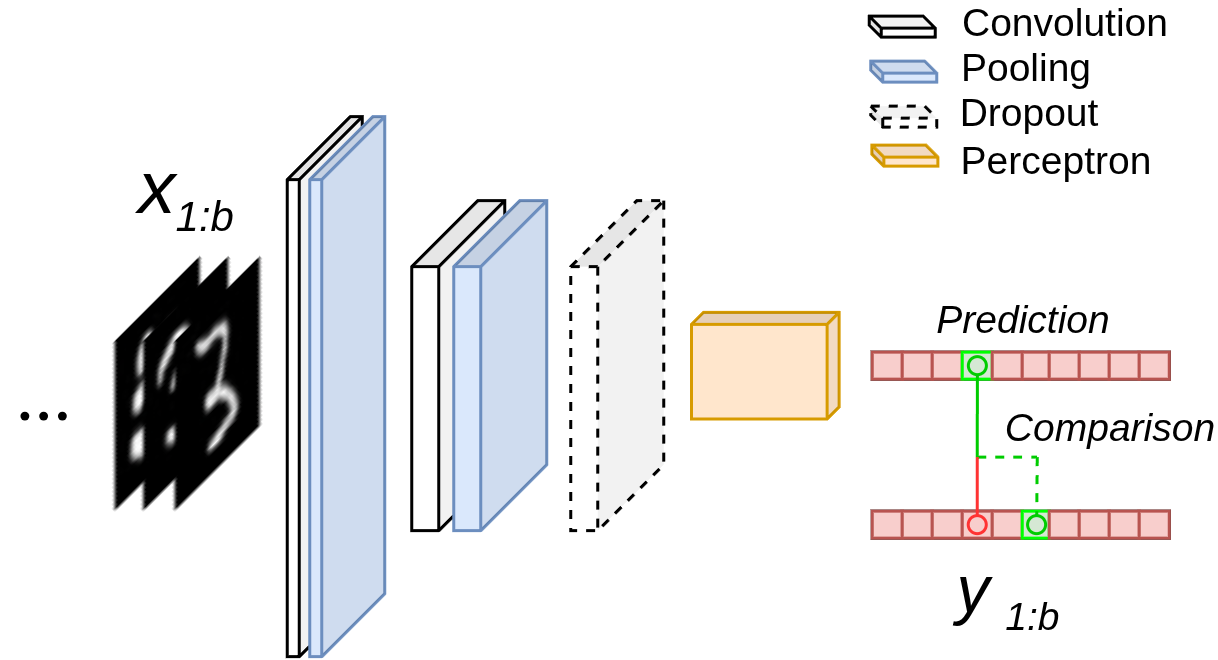}&
   %   } \hfill
     % \subfloat[Label Adulteration\label{fig:label_adulteration}]{
        \includegraphics[width=0.35\linewidth]{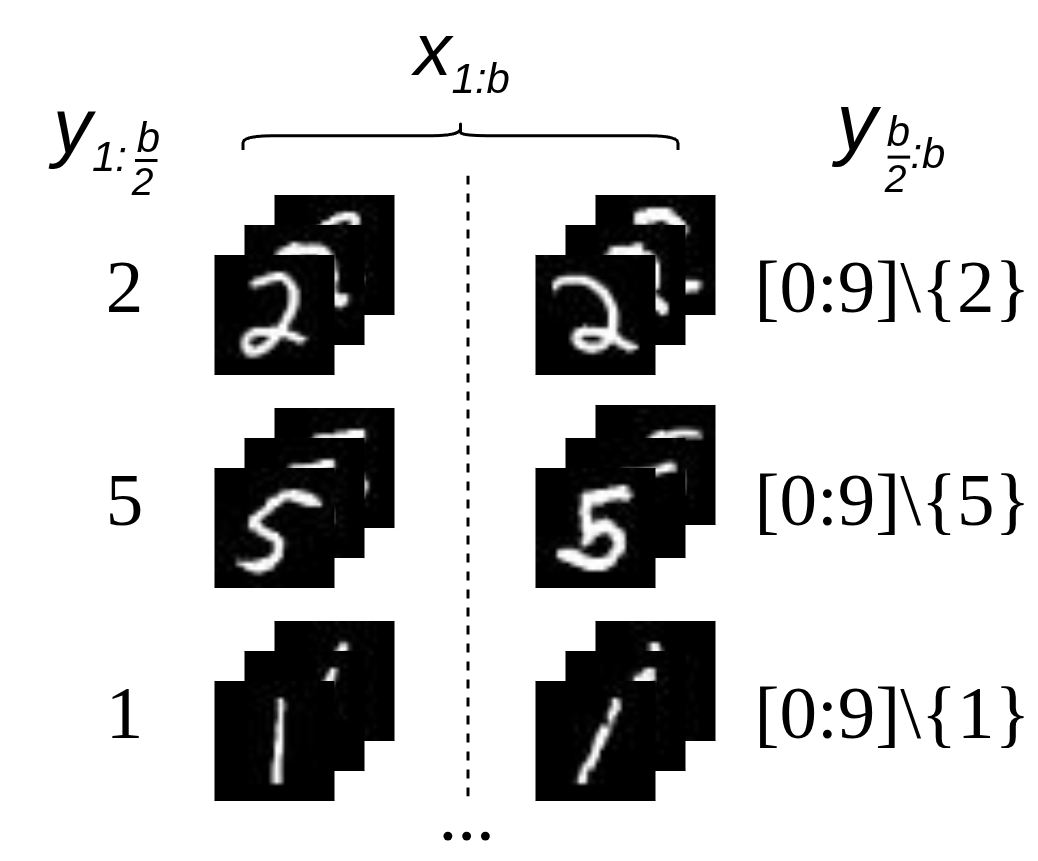}\\
     % }
     ({\bf a}) Task-oriented module &({\bf b}) Label Adulteration\\
\end{tabular}
      \caption{{(\textbf{a}) The VGG-like model trained and employed in the framework for a classification task and (\textbf{b})~$50\%$ label adulteration of the MNIST training data against respective visual content.}}
      \label{Fig:model_data}
    \end{figure}

    As for the second half of the MNIST dataset, it was used directly in the main surprise/pride experiments. This was adulterated so that $50\%$ of its instances had random labels, making them different from the specific digits they represented. The~adulteration is represented in \mbox{Figure~\ref{Fig:model_data}b}. Evidently, this was performed at random indices, so there would be good sparseness of correct and incorrect labels. Hence, despite being technically correct when classifying any of the $30,000$ images used in the experiment, the~pre-trained task-oriented network would be met with disparate labels approximately $50\%$ of the time. {This discrepancy is meant as the emotional trigger in the framework.}

\unskip

    \subsubsection{Actor--Critic~Module}
\begin{figure}[H]
        {\includegraphics[width=\columnwidth]{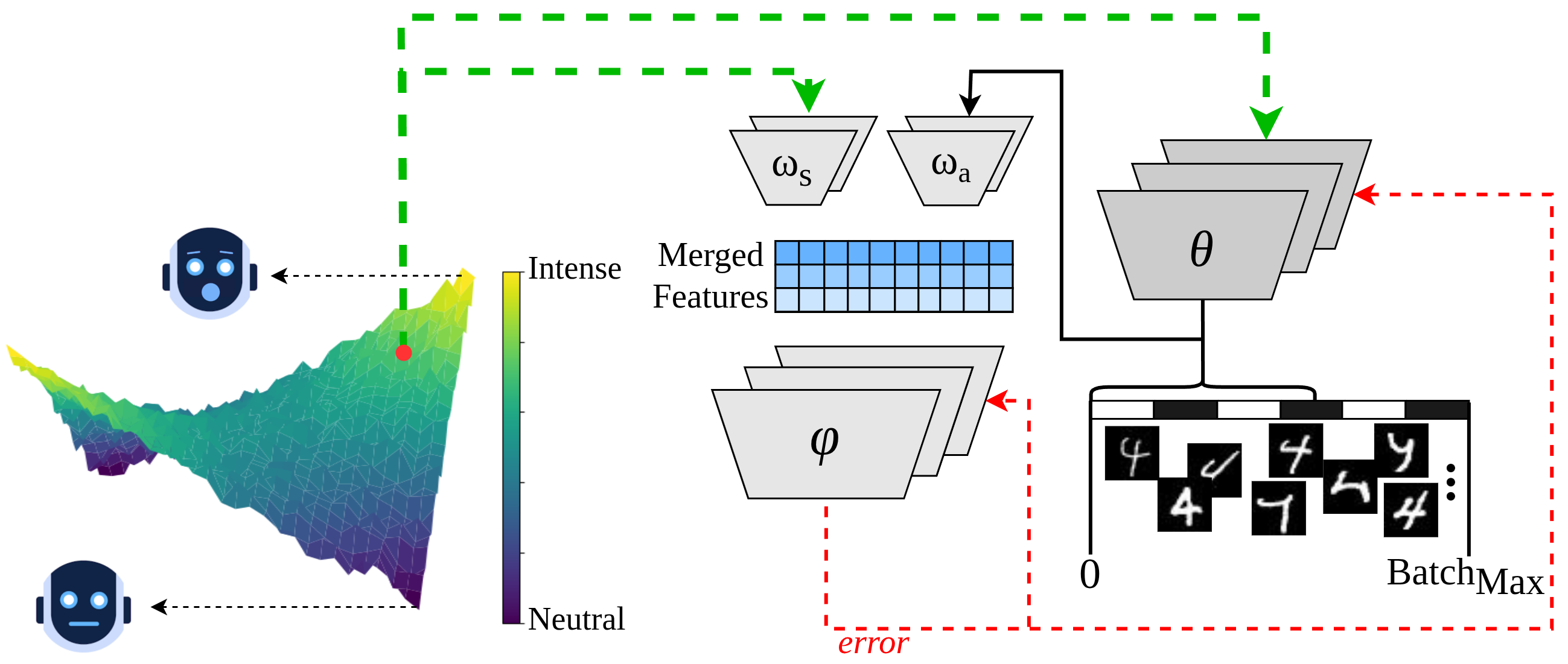}}
        \caption{{The actor--critic module. The~actor $\theta$ computes an exploratory rate based on received emotional scoring. The~critic $\omega - \phi$ scrutinizes this rate, generating feedback for itself and actor optimization towards task-oriented success.}}
        \label{Fig:actor_critic}
    \end{figure}

    Our system requires a continuous evaluation of its own emotional state, processing {it} into an exploratory rate, as~the most appropriate action. {To achieve this,} we {drew inspiration from} the habitual actor--critic dichotomy of the basal ganglia~\cite{O_Doherty_2004} {to design decision-making} neural modules in our artificial agents. {Since} we employ a form of directed exploration in our agent task, a~deterministic approach would be more fitting than a stochastic one. Hence, deep deterministic policy gradients (DDPGs)~\cite{LillicrapHPHETS15} were implemented as AI parallels of the basal ganglia. This type of reinforcement learning (RL) methodology assumes separate networks: the critic model and the actor model. {These collaborate to} map the state to the action deterministically, attempting to maximize reward. Both {the actor and the critic were implemented as} multi-layer perceptrons focused on generating embeddings, which are then reduced, respectively, as an action or rectifying signal. These embeddings are parsed from the emotional state of the artificial agent, which is taken as the sole input for the actor, and~tupled with the chosen action for the critic. An~overview of this process is shown in Figure~\ref{Fig:actor_critic}.

    {Within the RL paradigm,} the agent state is activated as formula-driven surprise or pride scores. {With it,} the actor decides on an exploratory rate {to be used by the task-oriented module.} The critic {then} signals the actor regarding that rate's task-oriented usefulness. This {translates as} attentional shifting, as~the actor {is effectively warranting} the task-oriented module to perform {more/less} intake of data in~order to mitigate/potentiate emotional exacerbation. {The resulting decision policy will therefore codify} the exploratory rate in terms of epistemic or achievement emotion. {We quantify this rate as a percentage to~extract a fraction of the batch size, which is bound by $[0, Batch_{Max}]$.} While not an exact representation of the basal ganglia and its related structures, this arrangement boasts similarities both architecturally and in terms of functioning of a human participant's decision-making process. {A comparison with human study results becomes more valid, in~line with RQ1.}

    %OLD CAPTION The module is composed of two separate neural models, for the actor and the critic respectively. Accuracy resulting from the task-oriented model is compounded with a confidence score, to compute an epistemic or achievement emotion, according to reports in cognitive psychology research. The actor $\theta$ receives this emotional score and decides on an exploratory rate for the task-oriented model. The critic model also receives this emotional score on its $\omega_s$ branch, in addition to the actor’s chosen exploration rate on its $\omega_a$ branch. The resulting merged features are processed by $\phi$ to generate a feedback signal on the actor’s decision and the critic’s own performance.

    \subsection{Learning~Cycle}
    Learning a correlation between surprise or pride and exploratory behavior involved all three models: {the task-oriented, actor, and critic models. Since the first is already trained and highly confident in its predictions, its weights are frozen and used for forward passes only. To~start,} the set of MNIST with partially adulterated labels {is} made available to the task-oriented model for~classification at instance-wise steps, which are mediated by the actor--critic module. {This means an} item is first picked randomly and processed to generate a corresponding label. The~adulteration of labels ensures that a portion of the predictions is unavoidably incorrect. Still, confidence remains high due to the initial training process {and weight freezing}. The~system is {now} able to experience both successful classification as well as incur high-confidence errors. Such circumstances induce emotional variation in accordance with the formulae described previously. This information is {inputted to} the actor, which will decide how much data of the same type should be analyzed subsequently {(i.e., explored), using the output rate to compute a batch fraction}. Processing of this {fraction} results in further emotional variation. {Its} comparison with single-instance states can yield insight into emotional progression during the cognitive task, in~addition to its relationship with exploratory fluctuation~overall.

    The emotional {scores} of the system and the actor-derived exploratory rate are taken in by the critic for adequacy assessments. This process {depends} on whether the chosen rate improves the system’s condition,~contributing towards its objective. {To specify this, we design a reward signal following the} standard assumption that participants typically intend to perform well in the activities they perform, maximizing success. Therefore, reward {varies} analogously to common human functioning wherein reward-coding neurons respond to success from profitable decision-making~\cite{Sirigu_2016} and maintenance of a homeostatic balance~\cite{Morville_2018}. {In our framework}, each RL agent obtains a basis reward {value}, whose polarity corresponds to that of the difference between explored batch accuracy and single-instance accuracy. {This ensures} that exploration {is} useful only if yielding improvements in terms of task performance. Additionally, {agents are} provided with a sparse reward matching the variation in epistemic/achievement emotion, which occurs during a step. This either minimizes surprise or maximizes pride, complying with the free energy principle, which illustrates a necessity of self-organizing agents to reduce uncertainty in future outcomes~\cite{Hartwig_2021}. The~reduction can stem from knowledge diversification, which is boosted by an exploratory increase, or~from near-complete reliance on current knowledge, where exploration is largely~avoided.

%%%%%%%%%%%%%%%%%%%%%%%%%%%%%%%%%%%%%%%%%%
\section{Experiments and Results}
The experimental application of our framework attempted to remain as close to Vogl's study as possible~\cite{Vogl_2019}. Since variations in nature/nurture naturally influence emotion and decision-making~\cite{Montag_2016}, {it is important to account for personality variability. Thus,} a total of 250 artificial agents were created, with~distinct emotional functions obtained through parameter variation and added noise. These were employed for surprise and pride experiments separately. {For either emotion function, $\mathcal{N}(0, 0.03)$ was employed, along with random combinations of $C_1$ and $C_2$ to generate varied artificial agents with individual differences while still following the same grand pattern.} The learning cycle was applied to each artificial agent, with~a reset occurring every 20 steps to match Vogl's 20-statement procedure. This reset marked the beginning of each learning episode, {with} a total of 100 episodes {solidifying} the robustness of our observations on emotion--exploration. Additionally, the~actor and critic models used the Adam optimizer during {the cycle run}, with~learning rates of 0.001 and 0.002, respectively. A~replay buffer was also implemented here to~reduce the variance from temporal correlations. Target networks {were also implemented to~help} regularize learning {updates}. Overall results are depicted in Figure~\ref{Fig:results}. {The} associations between exploratory behavior and epistemic/achievement emotions {are} analogous to findings reported in the original cognitive psychology study we strived to emulate~\cite{Vogl_2019}.

    \subsection{Outcome~Analysis}
\begin{figure}[H]
        {\includegraphics[width=\columnwidth]{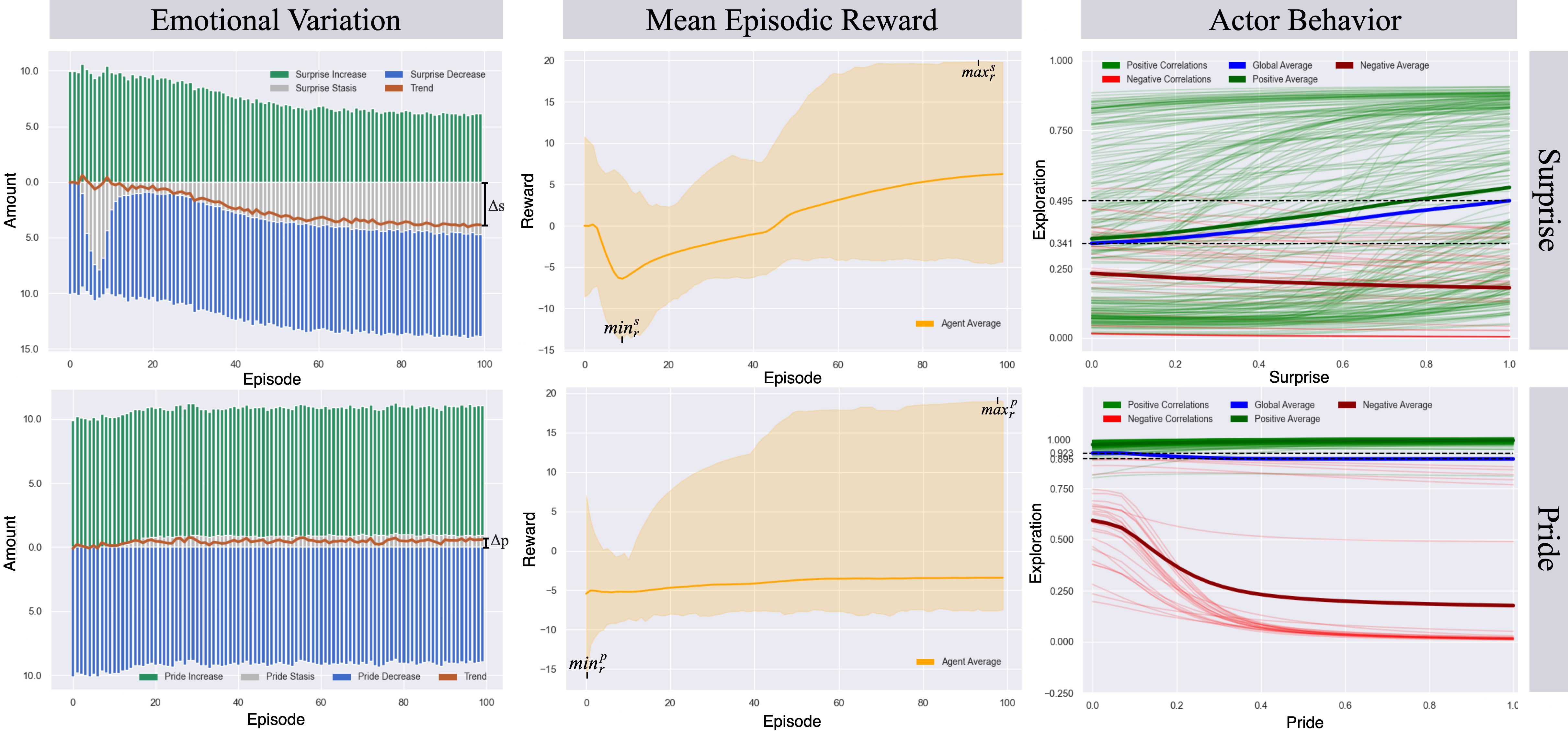}}
        \caption{{Results for both surprise and pride, mirroring similar findings in cognitive psychology. Leftmost column: Episodic mean of emotion differential between single sample and subsequent batch analysis steps across all implemented agents over the entire learning cycle. Middle column: Mean cumulative reward obtained by agents at each episode of the cycle. Rightmost column: Mean actor behavior at the end of the learning cycle, correlating surprise or pride with exploration.}}
        \label{Fig:results}
    \end{figure}

    First, model convergence was required to ensure behaviors learned by the artificial agents were not random. This was achieved {for either emotion, as~is demonstrated by the increase in and plateauing of} cumulative reward over time (Figure \ref{Fig:results} middle column). Specifically for surprise (top row), the initial reward is restricted to $[-8.58, 10.77]$, peaks at $max_{r}^{s} = 19.87$, and ends within the range of $[-4.37, 19.77]$, with~an early-stage dip minimum of $min_{r}^{s} = -13.64$. For~pride, the initial reward is encompassed by $[-15.97, 7.02]$, within~which $min_{r}^{p}$ is the minimum. The final reward here varies at $[-7.46, 18.95]$, though~the overall maximum value $max_{r}^{p} = 19.05$ is achieved shortly before. The average cumulative reward across agents also increases for both emotions. {This demonstrates} stable {yet} slight growth for pride. {Contrastingly, surprise entails} a short depression in earlier episodes followed by a steady increase {later on}. Overall, {these} trends indicate that agents successfully {learned} to correspond states to actions in a {useful} way.

    {As} episodic cumulative reward {validates} the success of the learning cycle, {observed} emotional fluctuations over time {are also legitimized. These are presented in the first column of} Figure~\ref{Fig:results}. {Expectedly}, the initial variation is well-balanced for both emotions, as~the number of increases matches that of decreases {in} the first episode. {However, different outcomes manifest as episodes progress:}
    \begin{itemize}
        \item Surprise (top trend): On average, bursts become less frequent by the final episode {($\Delta s = 38.52\%$). As~such, it seems a reduction or stasis is favored over increases.} Stasis is {also} progressively preferential in the first 10 episodes {of surprise, falling} back as decreases become more prominent later in the~cycle.

        \item Pride (bottom trend): {Overall,} the average emotion variation among agents is minor yet still favoring an upwards tendency. Decreases occur fewer times ($\Delta p = 5.90\%$ ) between the first and last episodes. {This percentage is largely taken over by stasis scenarios, with~the number of emotional increases remaining} mostly unchanged throughout the~cycle.
        %There is {also} no significant indicator of preference for increases over stasis.
    \end{itemize}

    \subsection{Observed~Correlations}
    {The third column of} Figure~\ref{Fig:results} evidences the impact {of emotion} over exploratory behavior. {In spite of} the emotional differences {imposed via} parameter fluctuation, a~substantial number of artificial agents {manifested} similar behaviors after the cycle. A~causation effect is most evident for the surprise experiment (top), which is {akin} to the emotional variation results. Averaging the decision-making behavior of all 250 agents displayed a $15.4\%$ increase in exploration in response to greater surprise. This trend {resonates with} the 217 agents {that} learned positive correlations, outshining the remainder 33 who {displayed} negative correlations. {Regardless of their} variability, all instances were monotonic and either displayed a considerable increase (positive) or a limited decrease (negative). {In the pride experiment (bottom),} instances were likewise monotonic. {However, agents here demonstrated} a deflating {exploratory} effect. Behaviors encompassed a large amount of positive weak correlations, {with} few yet ample negative correlations. Specifically, 222 agents displayed a slight exploratory increase with {pride. The impact} proved minimal, as~represented by the mean behavior of the subset. {A smaller set of 22 agents} decreased exploration between $25\%$ and $75\%$ towards null. {Contrastingly, this caused a substantial negative change in mean behavior. Moreover, it is aided by the six remaining agents who manifested} a more restrained reduction. {The final result is a} modest effect of a $2.8\%$ overall average decrease in exploration for increasing pride, despite{ several weak} positive correlations. {Here} medians can provide better insight {by} more accurately demonstrating negative cases as few yet hefty outliers against the~whole.

    %OLD CAPTION Results for both surprise and pride, mirroring similar findings in cognitive psychology. \textit{Leftmost column}: Episodic mean of emotion differential between single sample and subsequent batch analysis steps, across all implemented agents over the entire learning cycle. This shows a clear decrease for surprise (top) and slight increase for pride (bottom), which are biologically congruent. Moreover, this surprise minimization and pride maximization are in accordance with the free-energy principle, further supporting their validity. \textit{Middle column}: Mean cumulative reward obtained by agents at each episode of the cycle. While there is a clearer increase for surprise than for pride, both indicate agent convergence and learning of a useful relationship between surprise/pride and exploratory behavior. \textit{Rightmost column}: Mean actor behavior at the end of the learning cycle, correlating surprise or pride with exploration. A positive variation of exploratory behavior with surprise increase strongly resonates with Vogl’s experimental findings~\cite{Vogl_2019}, reporting the same behavior. As for pride, the obtained weak relationship further supports the conclusion that pride is not a strong precursor of exploration, as also demonstrated by Vogl’s diverging results for this emotion.

\begin{figure}[H]
        {\includegraphics[width=\columnwidth]{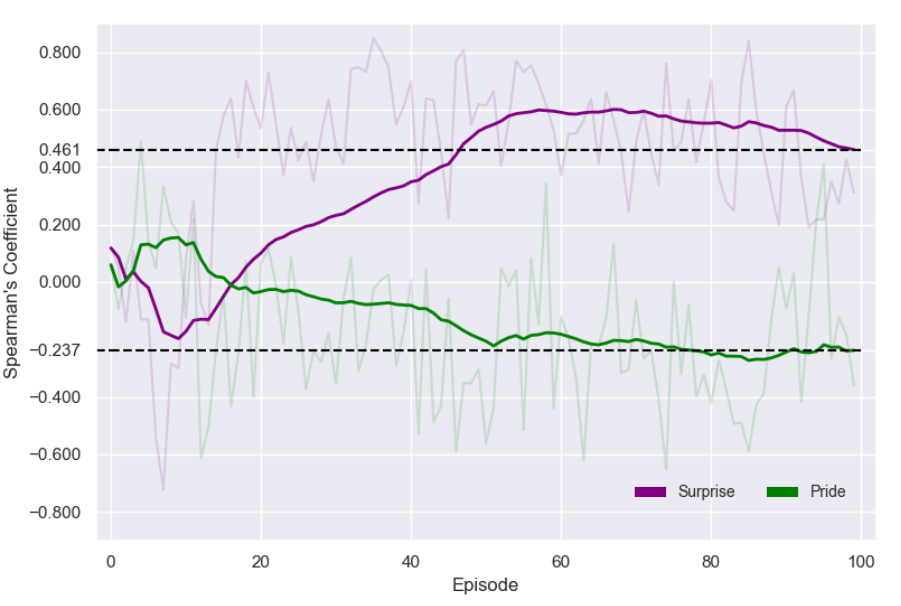}}
        \caption{{Agent episodic mean of Spearman's correlation coefficient between the actor-chosen exploratory rate and its causal surprise or pride score (pale), smoothed by a moving window of 40 samples (bold).}}
        \label{Fig:spearman}
    \end{figure}

    Relationship strength between exploration and pride/surprise can be further demonstrated by measuring how each data pair correlates throughout a cycle. {Here, we employ} Spearman's correlation coefficient $\rho$~\cite{spearman_co} {at} each episode in an agent's cycle. {This assesses if} observed monotonic relationships between emotion and exploration (Figure \ref{Fig:results} third column) become increasingly robust over time. {Resulting coefficients for either experiment, with} per-episode averaging of all 250 agent sample pairs, {are shown in} Figure~\ref{Fig:spearman}. {These} demonstrated considerable variability in the strength of the correlation between exploration and surprise/pride. {To mitigate this effect,} a sliding window encompassing 40 episodes was applied to smoothen the trends and clarify strength progression. Both {cases} display near-zero coefficients in earlier episodes. {However, this mean} $\overline{\rho}$ increases for surprise while decreasing for pride, resulting in $\rho_{surprise} = 0.461$ and $\rho_{pride} = -0.237$ by the end of the cycles. {This evidences} a moderate positive correlation for surprise and exploration. {For} pride, {there is instead a negative and much weaker} association with this behavior. {Succinctly, this weakness is congruent with the disparity of having} positive cases $10$ times more common than negatives ones {while still obtaining a negative correlation.}

    %OLD CAPTION Agent episodic mean of Spearman's correlation coefficient between actor-chosen exploratory rate and its causal surprise or pride score (pale), smoothed by a moving window of 40 samples (bold). As the learning cycle progresses, the positive relationship between surprise and exploration becomes more overt. Contrastingly, the negative relationship between this behavior and the emotion of pride does not progress as much, with the coefficient remaining closer to zero.

\section{Discussion}
{Results hold a clearly substantial} minimization of surprise. {This} conforms with the free energy principle, as~it led to a strong direct correlation being learned between this emotion {and} exploration. Contrarily, maximization of pride was somewhat negligible. This was paralleled by the weak dampening relationship obtained for pride over exploration, {with} its strength stagnating closer to zero. Notwithstanding the latter, both experiments successfully produced artificial agents capable of self-mediating their exploratory behavior. {They did so by} exploiting internal emotional drives towards improved task~performance. 

{The outcome of} our experiment strongly resonated with reports on emotion-mediated human exploratory behavior. Specifically, Vogl's study~\cite{Vogl_2019} demonstrated {a} causation effect {from} surprise over exploration of knowledge. {This was} evidenced by the successively positive path coefficients obtained when assessing surprise to curiosity and~curiosity to exploration effects. Additionally, {the} first and second versions of the study {reported} within-person correlation coefficients of $0.285$ {and} $0.262$ for this emotion and exploration. Even higher values {were reported when} considering curiosity intermediately. {Relating these findings with our work,} the $15.4\%$ mean exploratory increase leveraged by {the} agents over growing surprise validates {Vogl's} postulated path relationships. {Our} Spearman's correlation results {further support this, given the additional value proximity. For~surprise,} as the non-windowed coefficient mean across agents reaches $0.311$ by the final episode, {it becomes considerably} close to the within-person correlation value of the first study (with the most participants). {It also goes} reasonably near that of the second~study.

{In terms of pride, Vogl's} observations were {also} validated {by} our AI experiment. While negative correlation coefficients of $-0.073$ and $-0.177$ were {reported} in the first and second studies, respectively, these near-zero values indicate a weaker correlation between pride and exploration, if~any. {Though contradictory}, path coefficients first indicated a {weak but} positive causation effect {followed by a stronger} dampening impact later. {This suggested} that pride has a faint influence over exploration. {Our results are congruent, as~}Spearman's correlation {coefficient} took longer to deviate from null compared to the surprise experiment, {stagnating} at a lower absolute value. {Additionally, our} $2.8\%$ mean exploratory decrease over growing pride supports the higher likelihood of this emotion’s dampening effect over exploration. While this was due to a smaller amount of negative {strong} correlations, the~insignificance of a considerably larger amount of positive relationships {aligns with} pride's influence over exploration {being} modest. It could also be postulated that exploratory decrement is typical mostly during surges of {pride}. {This is because,} as a solo positive emotion, {it} may be damaging for cognitive performance~\cite{Bi_2022}, of~which exploratory behavior is a key aspect~\cite{Blanco_2016}. Additional experimentation would be required to test such hypotheses. {Regardless, under} equivalent test conditions, {our} agents {recognized a benefit in adopting behaviors akin to those exhibited} by humans. {These observations respond to RQ2, suggesting that emotion--exploration correlations manifested by agents are not only observable but also functionally identical. It is therefore plausible that these artificial correlations mirror their role in human cognition, being useful for data processing at least within comparable scenarios.}

{On a related note,} Vogl’s appealed for conceptual replication of their results to bolster generalizability~\cite{Vogl_2019, Vogl_2020}. {That is what inspired RQ1, to~which our proposed framework responds and meets the appeal. Also,} employing an image classification task is {evidently} different from the general knowledge trivia scenarios devised by those authors. {Thus, our results} further support the conclusion that observed emotion--exploration correlations were not {merely} triggered by {biased} input stimuli. {Moreover}, cognitive incongruity was induced differently, as~half of the data instances were assigned randomly incorrect labels at reset, rather than {always} being correct/incorrect. This meant contradictory information was a possibility, as~samples with visually distinct content could be assigned the same label. Vogl~et~al. also stressed the importance of considering various indicators of knowledge exploration {when attesting} the validity of an epistemic/achievement {source}. Unlike deep learning, where exploration may be {parameterized}, psychology relies on observations of behavior to assess {how exploration} occurs and to what extent~\cite{Gendolla_1997, Vogl_2019, Chevrier_2019}. As~{our} models objectively derive exploration from emotional scoring, this multi-indicator requisite was irrelevant in our approach, with~no impact over findings. Finally, implementing artificial agents as participants in a within-``person'' experiment is fundamentally distinct from human trialing. {This} constitutes further variability in comparison with {Vogl's} original~study.

\subsection*{Limitations}
{This work bears limitations, given the variability of its many factors and~the aim of interdisciplinary comparisons. For~instance, while authors strived to propose general solutions for emotional representation, the~functions may still introduce a potential bias which hinders generalizability. The~introduction of noise can help reduce this bias, yet a more robust approach would be to consider other functions that also match the psychological descriptions of surprise and pride and~to extend the experiment. A~similar limitation is related with the usage of MNIST and~classification as the cognitive task. We used a simplistic example, where high success rates are well-established in deep learning. Different outcomes could occur if considering a different task or~using a more complex dataset (e.g., CIFAR), and~thus bias may also exist. To~mitigate these limitations, we intend to vary the emotional functions further and~employ different datasets for classification so that~more generalizable conclusions can be made in future work. Still on the topic of validation,} our study is limited to 250~agents. {Naturally, experiment expansion can also entail a greater amount of agents in~order to} better reflect the diverse nature of living beings and their~behaviors.

{It is also important to note that} artificial agents are unlikely to bear conscious metacognition and be refrained to supporting humans in complex tasks in the foreseeable future~\cite{Korteling_2021}. {Thus,} a comparison of results obtained {from biological participants and AI} should be approached with care. {Given this limitation, we opt for the} corroborating usefulness {of agent observation. This can} provide further scrutiny for hypotheses on human cognition and behavioral traits~\cite{Wykowska_2016}. {Finally,} our system addresses each emotion individually as a precursor of exploration, whereas emotional states are most typically overlapping {and combine} effects over general behavior~\cite{Assuncao_2022}. {In the future, we intend to} expand the emotional basis of exploration to account for that overlap, further benefiting from and validating cognitive psychology hypotheses. For~instance, epistemic and achievement states could be employed in tandem, equitably, or via weighted contributions as~inputs to an actor module. {We speculate that emotional overlap is needed to stabilize correlations, causing more well-defined behaviors, similar to living beings. If~observed, this could provide further insights into human--AI parallelism and serve as inspiration for cognitive psychology expansion.}

%This technique could entail either feature merging at a midstream level, or an already multi-emotional input derived from a separate module. 

\section{Conclusions}
This work focused on developing a deep learning framework for emotional decision-making over exploration. The architectural design was inspired by basal ganglia circuitry, with~the foundation for emotional operation stemming from cognitive psychology. Emulation of human experimental conditions from {psichological studies} was conducted via an original learning cycle. {This} was applied to a {novel} deep learning framework, which was replicated {several} times for generalizability {purposes. This proposal adequately solves RQ1,} though others may still be possible. Furthermore, AI-learned correlations between epistemic/achievement states and exploration {demonstrated close proximity with} observations taken from human studies. {Hence, for~RQ2 we speculate} the correlations {to indeed} be useful for AI agents, much like they are to their human counterparts. {Our work additionally supports emotion-mediated learning, given its benefits to explainable AI and its autonomy. These include, for~instance, greater contextual adaptability for agents to self-adapt beyond just exploration or~a human-like understanding of AI decision-making.}

In terms of future research, our framework {follows the outlined benefits and others, since it can be adapted} for studying other behavioral traits and their relationship with emotional drives. Exploitation or engagement {may be mediated} through variable emotion during~cognitive operation. {This can} prove beneficial, similarly to how agents here learned to explore when it is seemingly more useful for their own reward objective. Finally, we speculate that further research on these topics {can} push AI closer towards {full} autonomy and general~intelligence.

%%%%%%%%%%%%%%%%%%%%%%%%%%%%%%%%%%%%%%%%%%
\vspace{6pt} 

%%%%%%%%%%%%%%%%%%%%%%%%%%%%%%%%%%%%%%%%%%
%% optional
%\supplementary{The following supporting information can be downloaded at:  \linksupplementary{s1}, Figure S1: title; Table S1: title; Video S1: title.}


% Only used for preprtints:
% \supplementary{The following supporting information can be downloaded at the website of this paper posted on \href{https://www.preprints.org/}{Preprints.org}.}

%\begin{tabularx}{\textwidth}{lll}
%\toprule
%\textbf{Name} & \textbf{Type} & \textbf{Description} \\
%\midrule
%S1 & Python script (.py) & Script of python source code used in XX \\
%S2 & Text (.txt) & Script of modelling code used to make Figure X \\
%S3 & Text (.txt) & Raw data from experiment X \\
%S4 & Video (.mp4) & Video demonstrating the hardware in use \\
%\bottomrule
%\end{tabularx}
%}

%%%%%%%%%%%%%%%%%%%%%%%%%%%%%%%%%%%%%%%%%%
\authorcontributions{Conceptualization, G.A., M.C.-B., and P.M.; methodology, G.A., M.C.-B., and P.M.; software, G.A.; validation, G.A.; data curation, G.A.; writing---original draft preparation, G.A.; writing---review and editing, M.C.-B. and P.M.; supervision, P.M. All authors have read and agreed to the published version of the~manuscript.}

\funding{This work has been financed by the PRR—Recovery and Resilience Plan—and by the Next-Generation EU European Funds, following NOTICE No. 02/C05-i01/2022, Component 5—Capitalization and Business Innovation (Mobilizing Agendas for Business Innovation)—under the project Greenauto (PPS10/PPS12/PPS13 with the reference 7255, C629367795-00464440).}

\institutionalreview{Not applicable.}

\informedconsent{Not applicable.}

\dataavailability{Publicly available datasets were analyzed in this study.  The~full MNIST dataset can be found in \url{https://www.kaggle.com/datasets/hojjatk/mnist-dataset}, accessed on 11 January 2023. No new data were created in this study.} 

\acknowledgments{This work was partially supported by FCT under grant 2020.05620.BD and OE (National funds of FCT/MCTES) under project UIDP/00048/2020.}

\conflictsofinterest{The authors declare no conflict of interest. The~funders had no role in the design of the study; in the collection, analyses, or~interpretation of data; in the writing of the manuscript, or~in the decision to publish the~results.} 

%%%%%%%%%%%%%%%%%%%%%%%%%%%%%%%%%%%%%%%%%%
%% Optional

%% Only for journal Encyclopedia
%\entrylink{The Link to this entry published on the encyclopedia platform.}

\abbreviations{Abbreviations}{
The following abbreviations are used in this manuscript:
\\

\noindent 
\begin{tabular}{@{}ll}
AI & Artificial Intelligence\\
CNN & Convolutional neural network\\
DDPGs & Deep Deterministic Policy Gradients\\
\end{tabular}
}

\begin{adjustwidth}{-\extralength}{0cm}
%\centering %% If there is a figure in wide page, please release command \centering, for Table, ``\textwidth" should be ``\fulllength"

\reftitle{References}

\PublishersNote{}
%\isPreprints{}{% This command is only used for ``preprints''.

\end{adjustwidth}
\end{document}